\documentclass{sig-alternate-05-2015}
\usepackage{hyperref}

\begin{document}

% Copyright
%\setcopyright{acmcopyright}
%\setcopyright{acmlicensed}
%\setcopyright{rightsretained}
%\setcopyright{usgov}
%\setcopyright{usgovmixed}
%\setcopyright{cagov}
%\setcopyright{cagovmixed}

% DOI
\doi{NA}

% ISBN
\isbn{NA}

%Conference
%\conferenceinfo{WWW 2017}{Submitted for review}

%\acmPrice{\$15.00}

%
% --- Author Metadata here ---
\conferenceinfo{Submitted to WWW}{'17 Perth, Australia}
%\CopyrightYear{2007} % Allows default copyright year (20XX) to be over-ridden - IF NEED BE.
%\crdata{0-12345-67-8/90/01}  % Allows default copyright data (0-89791-88-6/97/05) to be over-ridden - IF NEED BE.
% --- End of Author Metadata ---

\title{Combining observational and experimental data to find heterogeneous treatment effects}


\numberofauthors{2} %  in this sample file, there are a *total*
\author{
% 1st. author
\alignauthor
Alexander Peysakhovich\\
       \affaddr{Facebook Artificial Intelligence Research}\\
       \affaddr{770 Broadway}\\
       \affaddr{New York, NY 10003}
% 2nd. author
\alignauthor
Akos Lada\\
       \affaddr{Facebook News Feed}\\
       \affaddr{1 Hacker Way}\\
       \affaddr{Menlo Park, CA 94025}
}

\date{October 2016}
% Just remember to make sure that the TOTAL number of authors
% is the number that will appear on the first page PLUS the
% number that will appear in the \additionalauthors section.

\maketitle
\begin{abstract}
Every design choice will have different effects on different units. However traditional A/B tests are often underpowered to identify these heterogeneous effects. This is especially true when the set of unit-level attributes is high-dimensional and our priors are weak about which particular covariates are important. However, there are often observational data sets available that are orders of magnitude larger. We propose a method to combine these two data sources to estimate heterogeneous treatment effects. First, we use observational time series data to estimate a mapping from covariates to unit-level effects. These estimates are likely biased but under some conditions the bias preserves unit-level relative rank orderings. If these conditions hold, we only need sufficient experimental data to identify a monotonic, one-dimensional transformation from observationally predicted treatment effects to real treatment effects. This reduces power demands greatly and makes the detection of heterogeneous effects much easier. As an application, we show how our method can be used to improve Facebook page recommendations.
\end{abstract}

%\begin{CCSXML}


%
%  Use this command to print the description
%
\printccsdesc

% We no longer use \terms command


\section{Introduction}
Experimentation and data-grounded approaches to the design of products, websites and services have become immensely popular in the internet industry \cite{xu2015infrastructure, bakshy2014www}. This is for a very good reason: when decision-makers employ experimentation they have a far greater chance of making good decisions than with observation alone \cite{meyer2015two, banerjee2012poor, lalonde1986evaluating, kohavi2007practical, bottou2013counterfactual}. Standard experimentation techniques are often used to evaluate whether a particular policy works on average but decision-makers and analysts are increasingly interested in how their designs will affect different groups \cite{gail1985testing}. Understanding this heterogeneity is particularly important when a design is too expensive to give to the whole population (and so should be given to the subset that will benefit the most) or when it has positive effects for some, but may not be appropriate for others. Our contribution is to discuss how analysts can combine experimental with observational data to help solve the personalization problem.

Traditional methods have allowed analysts to specify ex-ante interesting subgroups \cite{gail1985testing, heckman2005structural} and look for differences. More modern work in machine learning, statistics and econometrics has begun to focus on streamlining this process. The methods automatically find heterogeneous groups using a variety of tricks such as non-parametric procedures \cite{crump2008nonparametric, taddy2014nonparametric}, regularized regression \cite{imai2013estimating}, trees \cite{green2012modeling, su2009subgroup}, causal trees and forests \cite{athey2016recursive, wager2015estimation}, deep neural networks \cite{shalit2016bounding}, `virtual twin' analyses \cite{foster2011subgroup} and mixtures of models \cite{grimmer2014estimating}.  

However, these automated searches are difficult. In many important domains average effects are small (relative to noise), the set of unit-level covariates is high-dimensional and we lack a priori knowledge about which covariates are important predictors of heterogeneity. The combination of these conditions means that to identify heterogeneous effects precisely experiments have to be very large and, in general, prohibitively costly \cite{lipsey1990design}. In contrast to expensive experimental data, observational data is often available in much larger quantities especially in applications such as medicine, online commerce or social media. Our contribution is to investigate whether we can combine this observational data with experiments to help learn the mapping from unit-level features to heterogeneous effects.

Our general approach can be described by the heuristic ``larger correlations suggest larger causal effects.'' We learn a mapping from unit-level covariates to the size of the unit-level causal effect that we get from using time series observational data. We then use this mapping as a rank-preserving transformation of the true causal effects. Of course our heuristic is not always applicable and we discuss the statistical assumptions on the data generating process under which the statement above is true.

Statistical assumptions are useful but often they are abstract and so analysts need guidelines for when they are likely to hold or not. In general, this requires domain knowledge and reasoning. We focus on a real world application: recommendations of pages to users on Facebook. We present a stylized model of user behavior and argue that under reasonable assumptions on this behavioral model the required statistical assumptions are satisfied. Of course, the final arbiter of such questions is data so we re-analyze an existing product A/B test to show that indeed observational data seems to provide a useful prior for experimentally estimated causal effects.

We note that our observational approach is not intended to supplant the approaches surveyed above. Rather, we view observational data as a complement to, not a substitute for, randomized trials. In practice, we suggest that the observationally predicted treatment effect for each unit be added as a feature into an analyst's favorite heterogeneous treatment effect procedure. 

There is a cost here: adding another feature raises the complexity of the machine learning procedure. There is also a benefit: if the conditions on the data generating processes we outline here are satisfied the benefits in predictive power may be large. We argue that if the covariate space is already large, the effect on model complexity is negligible \cite{friedman2001elements} and so the cost is likely to be small while the benefit in terms of reducing experimental power requirements may be quite large.

\section{The Basic Setup}
We work with the following situation: we have units indexed by $i$, time indexed by $t$, a continuous variable of interest $x$ and a continuous outcome variable $y$. We have a treatment which can change the value of $x$ by one and we are interested in personalizing the choice of whether a particular unit (i.e. group or individual) with a given covariate profile should or should not receive the treatment (perhaps the treatment has a cost or finite supply). 

To model this we assume each unit has a linear\footnote{We assume linearity because in many cases of interest our treatment will have relatively small effects on $x$ and thus we are interested in the locally linear approximation of the true response function.} response: when we raise $x_i$ by one, $y$ will increase by $\beta_i.$ To complete the model, we assume there is a (potentially high dimensional) space of covariates $C$ and each unit has a covariate profile $c_i$. Note that this covariate is assumed fixed per unit and not affected by policy choices.

There is a (potentially probabilistic) mapping $f$ from covariates to treatment effects $\beta_i = f(c_i)$ and the key to doing good personalization is to learn this mapping.

One way to estimate $f$ is to run a very large experiment. We can raise $x$ by one unit in treatment (leaving the control as is) and estimate the function $$\mathbb{E} (f(c_i)) = \mathbb{E} (y \mid treated, c_i) - \mathbb{E} (y \mid control, c_i)$$ using off-the-shelf methods \cite{taddy2014nonparametric, imai2013estimating, green2012modeling, su2009subgroup, athey2016recursive, wager2015estimation, shalit2016bounding, foster2011subgroup, grimmer2014estimating}. However, when treatment effects are relatively small and covariate space is large this technique will require quite large and expensive experiments.

On the other hand, highly granular observational data is often available in large quantities in online applications and increasingly also in the domains of medicine and public policy. We now discuss how to incorporate this observational data to help make the process of estimating heterogenous causal effects easier.

Suppose that the real data generating process is given by the structural equations $$x^{t}_i = \theta_i + \epsilon_{i}^t + U_{i}^{t} \psi_i$$ and $$y_i^t = \mu_{i} + x_{i}^t \beta_i + U_i^t \gamma_i + \eta_i^t.$$ Here we have that $\epsilon$ and $\eta$ are error terms independent of all the other variables, $U$ is some time varying unobserved variables (which in general can be a vector but for ease of notation we write as a scalar quantity), $\mu$ and $\theta$ are unobserved variables that are fixed at the unit level. For simplicity, all of these variables are mean $0$ with finite variances.

Note that this is without loss of generality as we could include some observed variables $V_i^t$, then the way to apply the method below is simply to use not the original $x$ and $y$ but $x$ and $y$ conditional on $V$ (ie. the residuals). For the rest of this paper we assume all variables other than $x$ and $y$ are unobserved by the analyst.

Suppose that for a large set of units $i$ we have observational data for $x$ and $y$ coming from a time series with periods $t \in \lbrace 1, \dots, T \rbrace$ where $T$ is large. Denote each daily data point for each unit as $(x^t_i, y^t_i)$. This is the standard panel data set up, however standard panel data techniques are generally interested in learning the average causal effect, not individualized ones \cite{angrist2008mostly}. 

What happens if we take this data and run a separate linear regression, including an intercept, for each unit? For simplicity we assume that $T$ is very large, so we can work with the population quantities free of uncertainty. First note that the panel aspect of our data means we can ignore $\mu_i$ and $\theta_i$ in the analyses that follow. This is because $\mathbb{E} (x) = \theta_i$ and $\mathbb{E}(y) = \mu_i + \beta_i \mu_i$. Both of these quantities are fixed \textit{within} units and thus picked up by the intercept term of each unit-specific linear regression.

Let us focus on the more interesting question: the coefficient on $x$ in each of our unit level regressions. We will call this our \textit{observationally estimated causal effect} $\hat{\beta}_i$. 

We don't observe $U$ and so our estimated $\hat{\beta}_i$ will not be equal to $\beta_i$ due to omitted variable bias \cite{angrist2008mostly, bottou2013counterfactual}. We abuse notation a bit and let $X_i$ be demeaned vector of observations of $x$ for unit $i$ (similarly for $Y_i$). Recall that the coefficient $\hat{\beta}_i$ is the solution to the least squares problem $$\hat{\beta}_i = (X'_i X_i)^{-1} (X'_i Y_i).$$ We can substitute the structural equation for $y$ (ignoring $\mu$ and $\theta$ because they have been picked up by removing the means) into the above to get  $$\mathbb{E} (\hat{\beta}_i) = \mathbb{E}((X'_i X_i)^{-1} (X'_i X_i \beta + U_i \gamma_i + \eta_i))$$ which after some algebra turns in to $$\mathbb{E}(\hat{\beta}_i) = \beta_i + \mathbb{E}((X'_i X_i)^{-1}(X'_i U_i) \gamma_i) = \beta_i + \frac{Cov(x_i, u_i)}{Var(x_i)} \gamma_i$$ where the last step comes from the fact that $x$ and $u$ are scalar.

This is an important example of the difference between prediction problems and interventional problems \cite{bottou2013counterfactual}, \cite{pearl2009causality}. $x_i\hat{\beta}_i$ is the \textit{best unbiased linear predictor} of $y_i$ but it is not an estimate of the causal effect $\beta$ for the same reason that knowing the number of units of ice cream sold on a given day can predict how many people will drown in swimming pools on that day but that that banning ice cream (an intervention) will likely not affect drownings by nearly as much as the observational regression would suggest. 

The bias term here is the quantity of interest for us. Suppose that we have two units $i$, $j$ and we have that $\beta_i > \beta_j.$ If $\frac{Cov(x_i, u_i)}{Var(x_i)} \gamma_i > \frac{Cov(x_j, u_j)}{Var(x_j)} \gamma_j$ then we are guaranteed to have $\hat{\beta}_i > \hat{\beta}_j.$ If this condition holds for any such $i$, $j$ then $\hat{\beta}$ derived from observational time series data will be a rank preserving estimate of $\beta.$ In other words, if ``bigger causal effects imply bigger bias'' then ``larger correlation effects imply larger causation.''\footnote{Note that this is a sufficient, but not necessarily condition. It is necessary that $\frac{\partial bias}{\partial \beta} > -1$ however this doesn't make for as catchy one sentence explanations or as simple to describe behavioral assumptions.}

In such a case $\hat{\beta}_i$ is a sufficient statistic for targeting budget constrained interventions (ie. when we can only afford to give the intervention to some percentage of the population). This also means that if we can learn a function $g$ which maps covariates $c_i$ to $\hat{\beta}_i$ then this function will be a monotonic transformation of the true heterogeneous effects function $f$. 

In practice, we suggest learning $g$ by estimating a set of $\lbrace \hat{\beta}_1, \dots, \hat{\beta}_N \rbrace$ for a large number of units using time series data and then using any standard machine learning technique to learn $\mathbb{E} (\hat{\beta} \mid c).$

Note that the existence of time series data is what make our job here possible because it allows us to remove time invariant unit-level effects. Without multiple observations per unit, it would theoretically be possible to perform a similar procedure but the assumptions required would be much stronger as we would need much more complex conditions on the covariance structure between $\mu$, $\theta$, $u$ and $\beta$ to hold.

Recall that our motivation is environments where observational data is plentiful but experimental data is sparse. Thus, if our assumptions above are satisfied, once we have $g$ we only need to use experimental data to learn a one dimensional transformation $h$ from $\hat{\beta}$ to $\beta$ rather than the full, high dimensional function $f$ (figure 1).

\begin{figure}[!ht]
	\begin{center}
		\includegraphics[scale=.25]{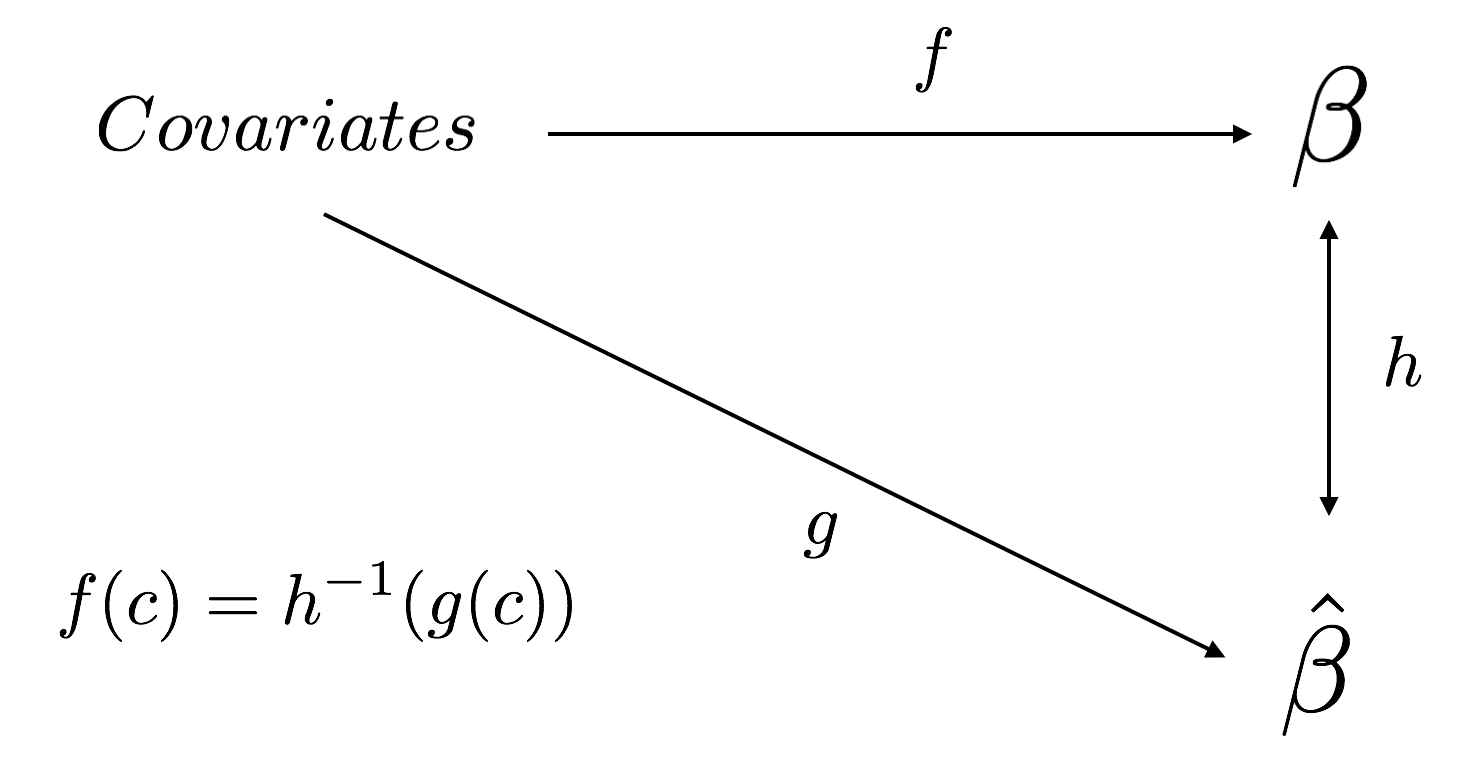}
	\end{center}
 	\caption{The relationship between the true heterogeneous treatment effect function $f$ and the observational one $g$.}
\end{figure}

There is no way to check whether the assumptions underlying our procedure hold or not using observational data only. However, we are not advocating that observational data replace experiments. When analyzing real results, analysts will likely already be using a technique to look for heterogeneous treatment effects using $C$ as a feature space. 

Our argument is simply that the analyst should add $\hat{g}(c)$ (the unit level predicted value of $\hat{\beta}$), to the set of unit-level features. In the worst case, this increases the complexity of the heterogeneous effect model slightly (assuming $C$ is already high dimensional). In the best case, our assumptions hold exactly, $\hat{g}(c)$ is a perfect monotonic transformation of $f(c)$ and thus using $\hat{g}(c)$ as a feature greatly reduces experimental power requirements.

\section{Application: Facebook Page Recommendations}
We now walk through an example where observational estimates can be helpful to analysts interested in understanding heterogeneous effects in the real world. We focus on personalization of Facebook's page recommendation engine. Pages are non-user entities on Facebook than can post content (e.g. news sites, blogs, certain celebrities). People can `Like' a page to connect to it. Liking a page makes posts that the page creates eligible to appear in the liker's News Feed. Helping users connect to content they care about can greatly improve their experience on the site, therefore Facebook employs recommendations to suggest relevant pages to users. These recommendations can show up in a user's News Feed in the form of `Pages You May Like' units (figure 2 for an example). 

Facilitating these connections has costs. There is opportunity cost (various `Recommended Page' units take up space on the News Feed) and there is a related user experience cost (users who do not want or need more connections can be inconvenienced or annoyed). Thus, unlike in the standard recommender system setting where the question of interest is \textit{which page} should we suggest \cite{bennett2007netflix}, an equally important question for us to address is: \textit{which users'} experience will be improved the most by additional page recommendations? 

\begin{figure}[!ht]
	\begin{center}
		\includegraphics[scale=.3]{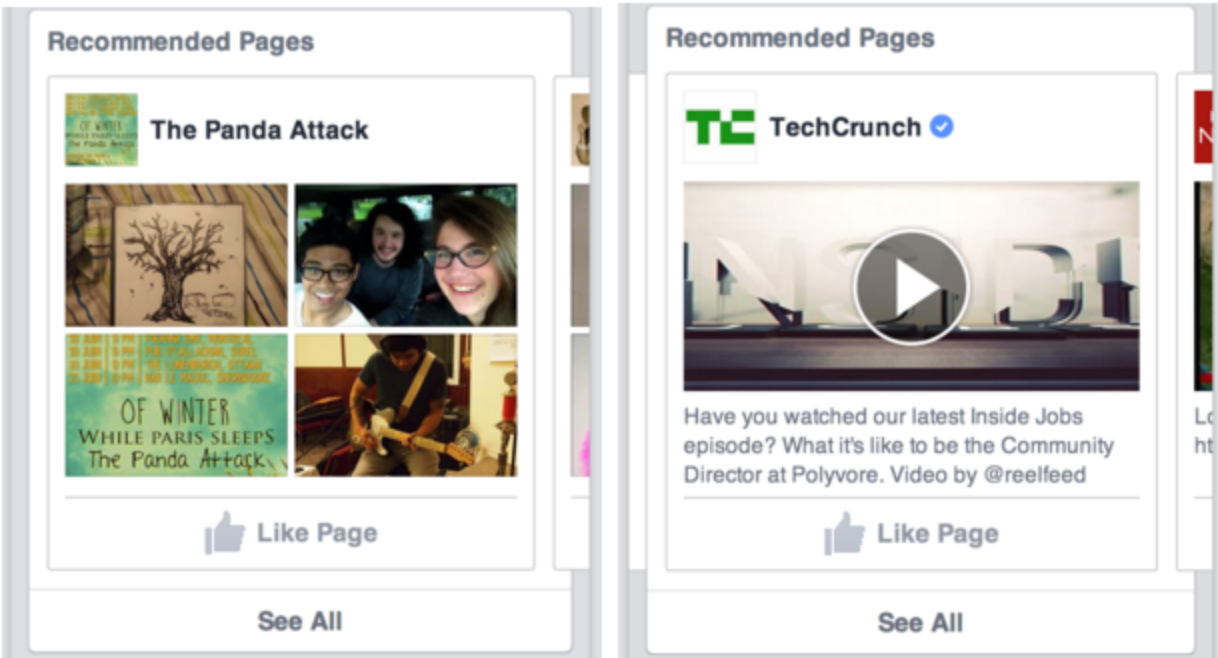}
	\end{center}
 	\caption{An example page recommendation unit on Facebook from 2014.}
\end{figure}

In the fall of 2015 Facebook tested a new type of recommendation - this unit would show a `representative` post from a page in a user's news Feed (with a header of `Recommended For You'). To construct these recommendations Facebook's standard recommender systems were used and only user-page pairs that had very high similarity scores according to this system were eligible for this new recommendation system. Other aspects of the recommendation unit's design were also fine tuned such that the recommendation was lightweight and did not detract from user experience.

To gauge whether this unit improved user experience, Facebook performed a standard A/B test on randomly chosen users. Approximately $8$ million people were eligible for this test (eligibility required that the underlying recommender systems were confident in their suggestions of potential new pages) and approximately $400,000$ were randomly chosen to see the new recommendation units. 

We will now show that in this example observationally estimated causal effects (from time series data) do indeed predict true causal effects estimated via an intervention.

\subsection{Behavioral Model}
First we discuss a microeconomic model of the world to get an intuition for whether the heuristic of ``larger correlations imply larger causal effects'' requires very stringent assumptions to hold or whether it is possible for this to be true in a stylized but realistic model. The ultimate judge of the effectiveness of our assumption will be data, but the model here is useful for gaining an intuition about our chances of success.

We define $x_i^t$ as a user's \textit{page inventory supply}. This is the number of posts that all pages that he or she is connected to (ie. has selected to be a fan of) have made on a given day. Note that not all of a users' page inventory is necessarily viewed or engaged with by that user.

For our outcome variable $y_i^t$, we will use overall time spent on Facebook by that user, on any desktop or mobile device. We will call this user $i$'s \text{demand for Facebook} on day $t$. We choose time spent as our measure of demand because there are many ways that a piece of page inventory could be interesting to a user: page inventory can be consumed passively (e.g. just by reading an article or watching a video). It can also lead a user to engage with similar content, the user can discuss the article in the comments section or in a group or reshare it to their friends. Asking whether a piece of inventory increases the total time spent on Facebook captures all of these various ways that piece of content can improve a user's experience.

Thus, we are interested in learning the mapping from $c$ to $\beta$ because individuals with large $\beta$ are ones who can benefit the most from additional page inventory in terms of increase their demand (as opposed to users who would not find page posts very interesting or those that already have more than enough inventory and would not benefit from any more). 

Let us consider a simple model for demand and supply. Suppose there is a single dimension on which individuals differ, which we'll call their affinity for Facebook denoted $a_i$. This dimension affects baseline demand for Facebook (because more active people have more friends and find more use from the platform) as well as their supply of page inventory (because more active people fan more pages). For clarity, we also assume that higher $a$ implies higher $\beta$. This would be true in a world where effects were different because latent heterogeneity is on a multiplicative scale. This dimension is latent and unobserved in our data directly. 

We also assume there is a time varying unobservable $e^t$ which is a property of a day. We call this the day's `event worthyness' - high $e$ days means that more things have happened on that day. This is another unobserved variable and we assume $e$ affects $x$ and $y$ as follows. First, higher $e$ days increase the demand $y$ through channels that are unrelated to page inventory supply (for example, by making it more attractive to come to Facebook to talk about the day's events either due to the individuals' own choices or due to the fact that more friends are posting and discussing). Second, higher $e$ affects page inventory supply because higher $e$ days cause pages to create more content.

Writing this in terms of our linear structural equations gives: $$x^{t}_i = D(a_i) + \epsilon_{i}^t + e^{t} \psi (a_i)$$ and $$y_i^t = \mu (a_i) + x_{i}^t \beta (a_i) + e^t \gamma (a_i) + \eta_i^t.$$ Running unit level regressions gives estimates $$\hat{\beta}(a_i) = \beta (a_i) + \gamma(a_i) \frac{Cov(x_i, e_i)}{Var(x_i)}$$ where the bias term can be rewritten by substituting the structural equation for $x$ as $$\gamma (a_i) \frac{\psi (a_i) \sigma_e}{\psi (a_i)^2 + \sigma_{\epsilon}}.$$ 

Since higher $a$ implies higher $\beta$ we also need the bias term to be increasing in $a.$ It seems reasonable to assume that in our model $\psi$ and $\gamma$ are both increasing in $a$. The former because we have made the assumption that more active people fan more pages (and thus their inventory supply is more affected by the occurrence of some event) and the latter because people with higher baseline affinity are more likely to have more friends and also to be more active on Facebook in general (thus events are more likely to generate additional demand for a user with high affinity than one with low affinity).

Therefore, to get the bias term to grow in $a$ we need that in general $\gamma (a) \psi(a)$ grows at a faster rate than $\psi (a)^2.$ Thus, we need $\gamma (a)$ to grow faster in $a$ than $\psi (a).$ This would imply in our case that the impact of affinity on baseline demand is larger than the impact of increasing affinity on page inventory supply. 

There are many things that affect Facebook demand (for example friends, groups joined, baseline like of Facebook, local community norms, quality of internet connection and page inventory supply). Higher $a$ likely increases all of these things and this total effect is what is captured in $\gamma$. On the other hand, $\psi$ captures only one piece of that sum. Thus, in our model the growth conditions above seem quite reasonable.

Again, we note that the purpose of this modeling exercise is to gain intuition, not claim that we've modeled the true data generating process. Rather by putting a stylized structure on the data generating process we can understand whether there is any chance that our statistical conditions hold and that our observationally estimated effect is related to our true causal effect. The ultimate arbiter will always be the data and that is what we turn to next.

\subsection{Estimating $\hat{\beta} (c)$}
We now estimate our observational causal effect. To do this, we take a random, deidentified, sample of 120 million Facebook users. We then compute $\hat{\beta}_i$ for each of the users by running the panel regressions as described above on 60 days of $(y^t_i, x^t_i)$ pairs with the definitions of $y$ and $x$ as in our behavioral model. This gives us our set of user-level coefficients $\hat{\beta}_i$. We estimate the function $\mathbb{E}(\hat{\beta} \mid c_i)$ using off the shelf machine learning.

Due to outliers, skew and noise in the estimates of $\hat{\beta}$ we change the regression problem to a classification problem. Instead of learning a predictor for continuous heterogenous treatment effects we perform a version of quantile regression \cite{angrist2008mostly} and give a user a label of 1 if their estimated $\hat{\beta_i}$ is in the top $20\%$ of all estimated $\hat{\beta}_i$ and $0$ otherwise. This also allows us to gauge our machine learning using AUC as a metric (which is more interpretable than MSE) as well as removes outliers (some estimated effects are very large or very small due to the fact that the first stage procedure is quite noisy).

For $c$ we take a large set of user-level covariates (these are standard covariates that analysts at Facebook have found useful including but not limited to average amounts of inventory of various kinds, engagements per content type, time spent, etc - the full set of features is too long to enumerate here) and use the Facebook machine learning stack which uses gradient boosted decision trees as feature transformers followed by a final linear layer (see \cite{he2014practical} for more details) to train a classifier using these features. 

The classifier gives us our function $\hat{g}$. Note that now the output of the classifier is the probability that a unit with covariate profile $c_i$ is in the top quantile of treatment effects. This is ok because we are not looking for actual numbers (because we don't believe that $\beta = \hat{\beta}$) but rather a rank ordering of units, which this probability provides.

\subsection{Testing $\hat{\beta}(c)$}
We now turn to verifying whether these observational estimates predict treatment effects in the product test described above. As our outcome measure we look at the total time spent over $1$ week of the experiment. To improve statistical power we transform the variable by taking natural logs and difference out $1$ week of pre-treatment time spent per unit (so our outcome variable is $\Delta y$ rather than just $y$). Because our outcome measure is so highly autocorrelated and right skewed these two transformations increase statistical power dramatically. We now ask: is there an increasing relationship between our learned $\mathbb{E}(\hat{\beta} \mid c)$ and the actual treatment effect in the experiment?

Figure 3 shows that the answer is yes. Stratifying the experimental population by predicted treatment effects shows an increasing relationship between $\hat{g}(c)$ and the actual treatment effect (the average difference in treatment and control). Note that the error bars are quite large because even though there are approximately $8$ million people in the control group, there are only approximately $400,000$ people in the treatment group and the experiment involves a very small change in the user experience.

We can also show this effect in a linear regression. We regress demand during the experimental period on an intercept, a treatment $0,1$ dummy, the predicted treatment effect, the interaction of the treatment dummy and the observationally predicted treatment effect and the pre-experiment unit-level demand (again this last term is just to reduce variance). We find a highly significant interaction (p<.01; thus confirming the visual impression of Figure 3). 

\begin{figure}[!ht]
	\begin{center}
		\includegraphics[scale=.65]{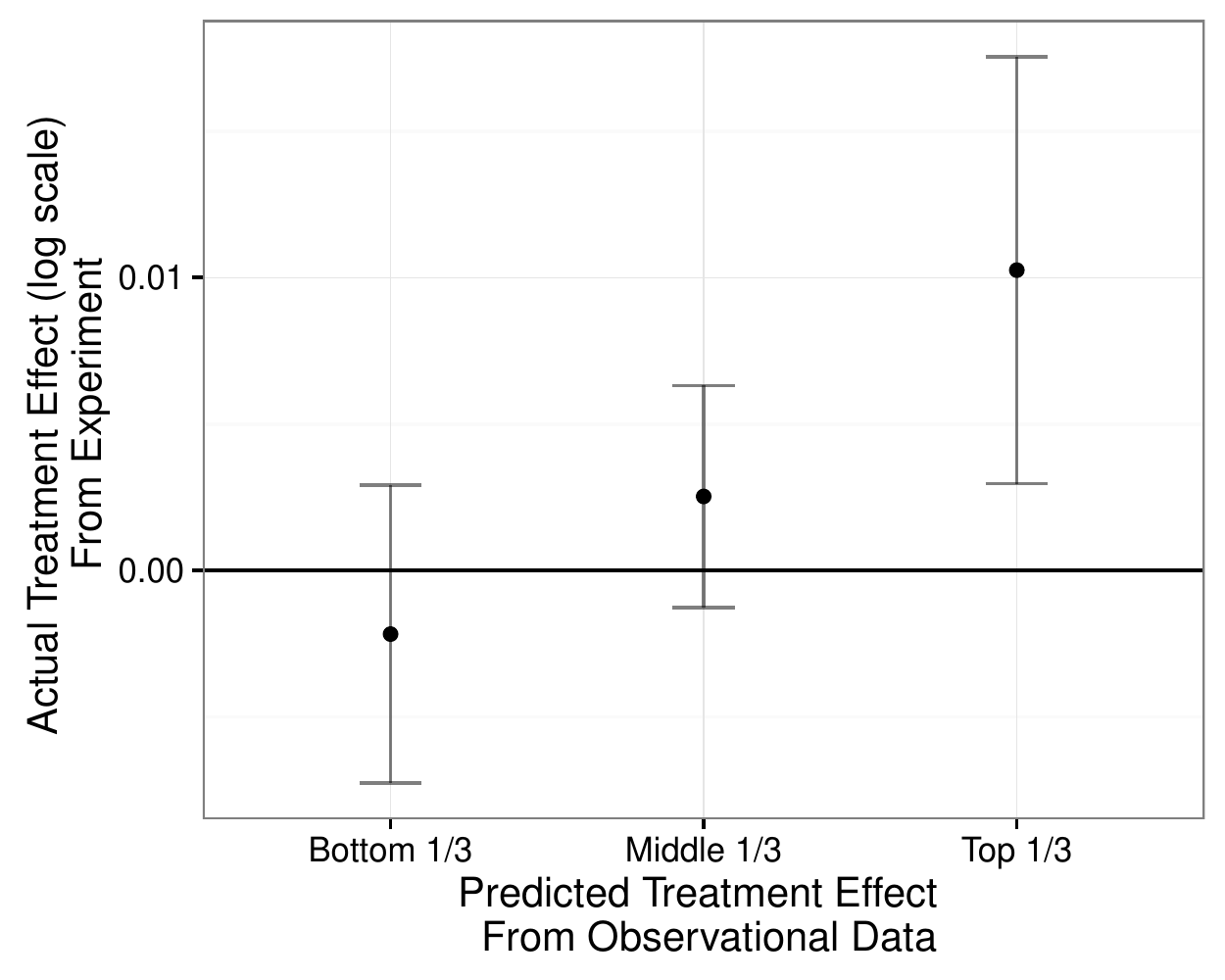}
	\end{center}
 	\caption{There is a monotonic relationship between predicted treatment effects and actual treatment effects.}
\end{figure}

\section{Conclusion}
Estimating heterogeneous effects is an important but very difficult problem. It is particularly difficult when the set of unit-level covariates is large and priors about which ones are important are weak. We have shown a method that uses observational data to get an estimate for true heterogeneity in causal effects. Our method requires assumptions on the data generating process, but we have discussed how analysts can think through whether these assumptions are likely to be satisfied in practice. In addition, we have argued that the cost of including an additional variable (the observationally predicted treatment effect) in experimental analyses is likely small but the gain is potentially large. We have shown that this is true on a real world task.

There are many possible extensions to our method. From the applied side we think that incorporating observational and experimental data into a single end-to-end training procedure (rather than step-wise as we do) is an interesting direction for future work. From the statistical side we have made strong assumptions about the data generating process to get our results. However whether there are weaker assumptions which yield weaker but nonetheless useful results about the relationship between observationally and experimentally estimated treatment effects remains open.

The method presented here bears conceptual resemblance to recent work on combining survey results from a probability sample and a biased sample using `data enriched' linear regression \cite{chen2015data}. It seems to us that both our work here and the results discussed there fit into the much larger growing literature on semisupervised learning \cite{zhu2005semi}, \cite{collobert2008unified} or multitask learning \cite{caruana1998multitask}. The idea behind such approaches is that machine learning can be made more efficient by performing multiple tasks at the same time.

The intuition behind why this can occur is that representations of the data learned from doing one task can be useful for other related tasks (for example the embedding of images learned from classifying whether a picture contains a cat may also be useful to detect whether an picture contains a dog). Viewed in this framework, our method uses the observational task to represents each unit by a scalar (the observationally estimated causal effect), However, we could also imagine training the map from $c$ to $\hat{\beta}$ using a neural network and, instead of using the predictions as the feature of interest for our experimental heterogeneous effects classifier we could use some intermediate representation of the unit features learned from the neural network. This seems like a promising avenue for future work.

It's hard to make good decisions without data and recent years have seen an explosion in the scope and complexity of data that decision-makers can use. However, data by itself is useless without methods to process it. We hope that our research contributes to both the theoretical and applied aspects of this vital discussion.

\bibliographystyle{abbrv}
\bibliography{regressionator}

\begin{thebibliography}{10}

\bibitem{angrist2008mostly}
J.~D. Angrist and J.-S. Pischke.
\newblock {\em Mostly harmless econometrics: An empiricist's companion}.
\newblock Princeton university press, 2008.

\bibitem{athey2016recursive}
S.~Athey and G.~Imbens.
\newblock Recursive partitioning for heterogeneous causal effects.
\newblock {\em Proceedings of the National Academy of Sciences},
  113(27):7353--7360, 2016.

\bibitem{bakshy2014www}
E.~Bakshy, D.~Eckles, and M.~S. Bernstein.
\newblock Designing and deploying online field experiments.
\newblock In {\em Proceedings of the 23rd ACM conference on the World Wide
  Web}. ACM, 2014.

\bibitem{banerjee2012poor}
A.~Banerjee and E.~Duflo.
\newblock {\em Poor economics: A radical rethinking of the way to fight global
  poverty}.
\newblock PublicAffairs, 2012.

\bibitem{bennett2007netflix}
J.~Bennett and S.~Lanning.
\newblock The netflix prize.
\newblock In {\em Proceedings of KDD cup and workshop}, volume 2007, page~35,
  2007.

\bibitem{bottou2013counterfactual}
L.~Bottou, J.~Peters, J.~Q. Candela, D.~X. Charles, M.~Chickering,
  E.~Portugaly, D.~Ray, P.~Y. Simard, and E.~Snelson.
\newblock Counterfactual reasoning and learning systems: the example of
  computational advertising.
\newblock {\em Journal of Machine Learning Research}, 14(1):3207--3260, 2013.

\bibitem{caruana1998multitask}
R.~Caruana.
\newblock Multitask learning.
\newblock In {\em Learning to learn}, pages 95--133. Springer, 1998.

\bibitem{chen2015data}
A.~Chen, A.~B. Owen, M.~Shi, et~al.
\newblock Data enriched linear regression.
\newblock {\em Electronic Journal of Statistics}, 9(1):1078--1112, 2015.

\bibitem{collobert2008unified}
R.~Collobert and J.~Weston.
\newblock A unified architecture for natural language processing: Deep neural
  networks with multitask learning.
\newblock In {\em Proceedings of the 25th international conference on Machine
  learning}, pages 160--167. ACM, 2008.

\bibitem{crump2008nonparametric}
R.~K. Crump, V.~J. Hotz, G.~W. Imbens, and O.~A. Mitnik.
\newblock Nonparametric tests for treatment effect heterogeneity.
\newblock {\em The Review of Economics and Statistics}, 90(3):389--405, 2008.

\bibitem{foster2011subgroup}
J.~C. Foster, J.~M. Taylor, and S.~J. Ruberg.
\newblock Subgroup identification from randomized clinical trial data.
\newblock {\em Statistics in medicine}, 30(24):2867--2880, 2011.

\bibitem{friedman2001elements}
J.~Friedman, T.~Hastie, and R.~Tibshirani.
\newblock {\em The elements of statistical learning}, volume~1.
\newblock Springer series in statistics Springer, Berlin, 2001.

\bibitem{gail1985testing}
M.~Gail and R.~Simon.
\newblock Testing for qualitative interactions between treatment effects and
  patient subsets.
\newblock {\em Biometrics}, pages 361--372, 1985.

\bibitem{green2012modeling}
D.~P. Green and H.~L. Kern.
\newblock Modeling heterogeneous treatment effects in survey experiments with
  bayesian additive regression trees.
\newblock {\em Public Opinion Quarterly}, page nfs036, 2012.

\bibitem{grimmer2014estimating}
J.~Grimmer, S.~Messing, and S.~J. Westwood.
\newblock Estimating heterogeneous treatment effects and the effects of
  heterogeneous treatments with ensemble methods.
\newblock {\em Unpublished manuscript, Stanford University, Stanford, CA},
  2014.

\bibitem{he2014practical}
X.~He, J.~Pan, O.~Jin, T.~Xu, B.~Liu, T.~Xu, Y.~Shi, A.~Atallah, R.~Herbrich,
  S.~Bowers, et~al.
\newblock Practical lessons from predicting clicks on ads at facebook.
\newblock In {\em Proceedings of the Eighth International Workshop on Data
  Mining for Online Advertising}, pages 1--9. ACM, 2014.

\bibitem{heckman2005structural}
J.~J. Heckman and E.~Vytlacil.
\newblock Structural equations, treatment effects, and econometric policy
  evaluation.
\newblock {\em Econometrica}, 73(3):669--738, 2005.

\bibitem{imai2013estimating}
K.~Imai, M.~Ratkovic, et~al.
\newblock Estimating treatment effect heterogeneity in randomized program
  evaluation.
\newblock {\em The Annals of Applied Statistics}, 7(1):443--470, 2013.

\bibitem{kohavi2007practical}
R.~Kohavi, R.~M. Henne, and D.~Sommerfield.
\newblock Practical guide to controlled experiments on the web: listen to your
  customers not to the hippo.
\newblock In {\em Proceedings of the 13th ACM SIGKDD international conference
  on Knowledge discovery and data mining}, pages 959--967. ACM, 2007.

\bibitem{lalonde1986evaluating}
R.~J. LaLonde.
\newblock Evaluating the econometric evaluations of training programs with
  experimental data.
\newblock {\em The American economic review}, pages 604--620, 1986.

\bibitem{lipsey1990design}
M.~W. Lipsey.
\newblock {\em Design sensitivity: Statistical power for experimental
  research}, volume~19.
\newblock Sage, 1990.

\bibitem{meyer2015two}
M.~N. Meyer.
\newblock Two cheers for corporate experimentation: The a/b illusion and the
  virtues of data-driven innovation.
\newblock {\em J. on Telecomm. \& High Tech. L.}, 13:273, 2015.

\bibitem{pearl2009causality}
J.~Pearl.
\newblock {\em Causality}.
\newblock Cambridge university press, 2009.

\bibitem{shalit2016bounding}
U.~Shalit, F.~Johansson, and D.~Sontag.
\newblock Bounding and minimizing counterfactual error.
\newblock {\em arXiv preprint arXiv:1606.03976}, 2016.

\bibitem{su2009subgroup}
X.~Su, C.-L. Tsai, H.~Wang, D.~M. Nickerson, and B.~Li.
\newblock Subgroup analysis via recursive partitioning.
\newblock {\em The Journal of Machine Learning Research}, 10:141--158, 2009.

\bibitem{taddy2014nonparametric}
M.~Taddy, M.~Gardner, L.~Chen, and D.~Draper.
\newblock A nonparametric bayesian analysis of heterogeneous treatment effects
  in digital experimentation.
\newblock {\em arXiv preprint arXiv:1412.8563}, 2014.

\bibitem{wager2015estimation}
S.~Wager and S.~Athey.
\newblock Estimation and inference of heterogeneous treatment effects using
  random forests.
\newblock {\em arXiv preprint arXiv:1510.04342}, 2015.

\bibitem{xu2015infrastructure}
Y.~Xu, N.~Chen, A.~Fernandez, O.~Sinno, and A.~Bhasin.
\newblock From infrastructure to culture: A/b testing challenges in large scale
  social networks.
\newblock In {\em Proceedings of the 21th ACM SIGKDD International Conference
  on Knowledge Discovery and Data Mining}, pages 2227--2236. ACM, 2015.

\bibitem{zhu2005semi}
X.~Zhu.
\newblock Semi-supervised learning literature survey.
\newblock 2005.

\end{thebibliography}
  % sigproc.bib is the name of the Bibliography in this case
% ACM needs 'a single self-contained file'!
%\balancecolumns
\end{document}